\documentclass[letterpaper, 10 pt, conference]{ieeeconf}  

\IEEEoverridecommandlockouts                              

\usepackage{amsmath} 
\usepackage{amsfonts}
\usepackage{todonotes}
\usepackage{url}
\let\labelindent\relax  
\usepackage{enumitem} 
\usepackage{multirow}
\usepackage{verbatim}
\usepackage{flushend}

\usepackage{tikz}		
\usepackage{textcomp}
\usepackage{hyperref}
\usepackage{lipsum}

\title{\LARGE \bf
A Comparison of LiDAR-based SLAM Systems for Control of Unmanned Aerial Vehicles
}

\author{Robert Milijas, Lovro Markovic, Antun Ivanovic, Frano Petric and Stjepan Bogdan
	\thanks{Authors are with Faculty of Electrical and Computer Engineering,
        University of Zagreb, 10000 Zagreb, Croatia
        {\tt\small (robert.milijas, lovro.markovic, antun.ivanovic, frano.petric, stjepan.bogdan) at fer.hr}}}%

\newcommand\copyrighttext{%
	\footnotesize \textcopyright 2021 IEEE. Personal use of this material is permitted. Permission from IEEE must be obtained for all other uses, in any current or future media, including 
	reprinting/republishing this material for advertising or promotional purposes, creating new 
	collective works, for resale or redistribution to servers or lists, or reuse of any copyrighted 
	component of this work in other works. 
	DOI: \href{https://doi.org/10.1109/ICUAS51884.2021.9476802}{10.1109/ICUAS51884.2021.9476802}}
\newcommand\copyrightnotice{%
	\begin{tikzpicture}[remember picture,overlay]
	\node[anchor=south,yshift=10pt] at (current page.south) {\fbox{\parbox{\dimexpr\textwidth-\fboxsep-\fboxrule\relax}{\copyrighttext}}};
	\end{tikzpicture}%
}

\begin{document}

\maketitle
\copyrightnotice
\thispagestyle{empty}
\pagestyle{empty}

\begin{abstract}

This paper investigates the use of LiDAR SLAM as a pose feedback for autonomous flight. Cartographer, LOAM and HDL graph SLAM are first introduced on a conceptual level and later tested for this role. They are first compared offline on a series of datasets to see if they are capable of producing high-quality pose estimates in agile and long-range flight scenarios.
The second stage of testing consists of integrating the SLAM algorithms into a cascade PID UAV control system and comparing the control system performance on step excitation signals and helical trajectories. The comparison is based on step response characteristics and several time integral performance criteria as well as the RMS error between planned and executed trajectory.

\end{abstract}

\section{Introduction} \label{sec:introduction}

The world of Unmanned Aerial Vehicles (UAVs) is rapidly evolving. Relatively simple design, well established control algorithms, and an affordable price range make such vehicles the preferred choice of professionals, hobbyists and researchers. The former two groups mostly rely on piloting skills when employing UAVs. On the other hand, the research community is constantly investigating new approaches and expanding UAVs' autonomous abilities which make UAVs versatile in a wide range of environments.

Reliable pose sensing plays a crucial role for UAVs as it is required for navigation and control of such vehicles. Motion capture systems, such as \textit{OptiTrack} and \textit{Vicon}, are suitable for providing  high-precision pose estimates. However, such systems are restricted to relatively small indoor laboratory environments. Outdoor environments do not have the luxury of employing such complex and sensitive systems. Satellite navigation through the Global Navigation Satellite System (GNSS) or similar systems can provide reliable measurements when operating in open areas or if paired with Real-Time Kinematic (RTK) positioning \cite{ferreira2020}. However, if the mission considers navigating through a GNSS denied environment, such systems are not sufficient.


Simultaneous Localization and Mapping (SLAM) algorithms are capable of localizing based only on measurements from the onboard sensory apparatus. In the field of aerial robotics, two different approaches are dominantly used: visual SLAM with monocular or stereo cameras; and Light Detection and Ranging (LiDAR) sensors. The visual SLAM methods employ cameras and image processing algorithms \cite{Taketomi2017}. Researchers in \cite{Perez-Grau2017} are focused on planning obstacle free trajectories for a UAV in GPS denied environments, based on a RGB-D SLAM. The work in \cite{Lin2018} uses a monocular fish-eye camera and an Inertial Measurement Unit (IMU) as a minimal sensor suite for performing localization. In \cite{rizik2020} an image stitching technique is applied in SLAM for autonomous UAV navigation in GNSS denied environments. Moreover, researchers in \cite{bavle2020} developed a Visual Planar Semantic (VPS) graph SLAM method that uses a neural network to detect objects, while the work in \cite{martinez2015} considers ORB-SLAM for UAVs.

\begin{figure}[t]
	\centering
	\includegraphics[trim= 0 50 0 110, clip, width=0.99\linewidth]{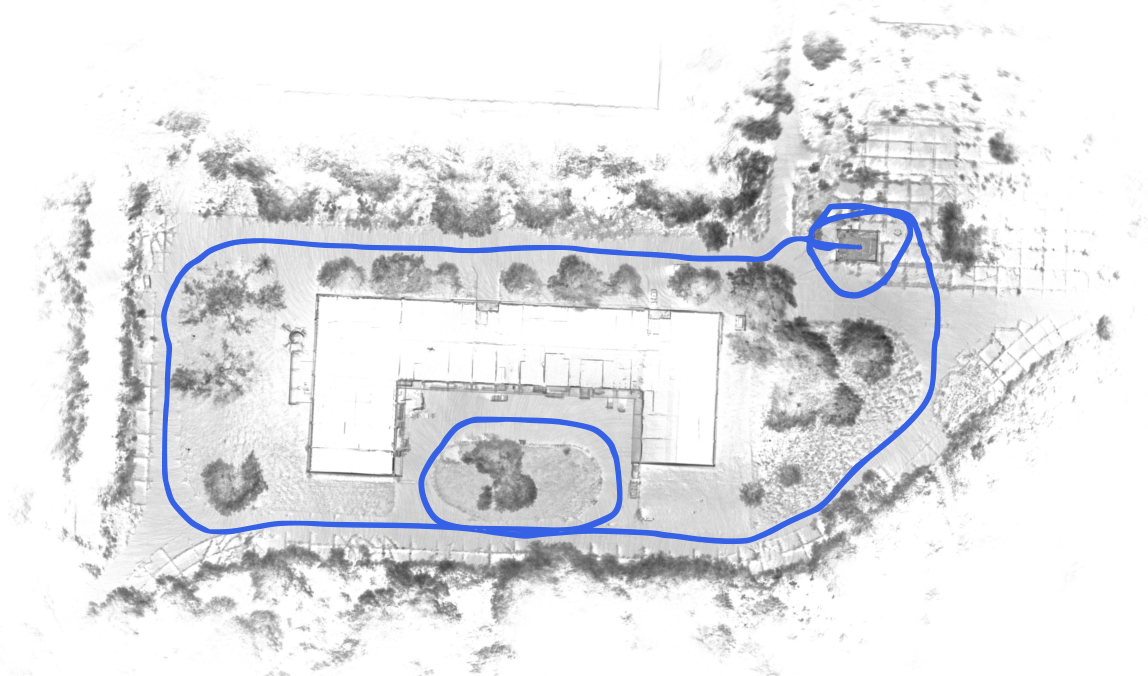}
	\caption{Top-down view of a map created by the Cartographer SLAM algorithm. The blue line denotes the UAV trajectory. The size of the mapped environment is roughly $250m \times 100m$.}
	\label{fig:preliminary_3_carto_points}
\end{figure}

Although cameras are an excellent choice for UAVs due to their small weight and size, LiDAR sensors are also used for SLAM. Fig. \ref{fig:preliminary_3_carto_points} is an example of a map created based on data received from an airborne 3D LiDAR. The main advantage of 3D LiDARS is providing a point cloud directly, opposed to cameras where some image processing techniques have to be employed to obtain the point cloud. As technology progressed, size and weight of LiDARs has come down, which places them within the UAV payload capabilities. For example, \cite{Lopez2016} uses both a 2D LiDAR and a monocular camera to build 2.5D maps with a micro aerial vehicle. In \cite{Kumar2017} two 2D LiDAR sensors are mounted perpendicularly on the UAV to obtain a 3D map of an underground environment.


Some examples of 3D LiDAR usage include multiple sensors. The work in \cite{Paneque_2019} fuses stereo camera, 3D LiDAR, altimeter and Ultra-Wideband Time-Of-Flight (UWB TOF) sensor measurements for localization within a known map. \cite{Ozaslan_2017} fuses information from an IMU, a 3D LiDAR, and three cameras to achieve autonomous inspection of penstocks and tunnels with UAVs. In \cite{Alexis_2019} and \cite{Khattak_2020}, researchers use a fusion of LiDAR Odometry And Mapping (LOAM) with IMU, thermal odometry and visual odometry to navigate UAVs in underground mines.


The use of LOAM in fusion with other sensors has sparked our interest to use a LiDAR SLAM algorithm on its own for aerial pose sensing. 
As a result, the focus of this paper is comparison between contemporary SLAM algorithms for that purpose. 
Algorithms will be compared based on the following criteria:
\begin{enumerate}[label=(\roman*)]
	\item Map quality in three scenarios - small map, dynamically challenging maneuvers, and large map;
	\item Feedback and control quality in two scenarios: step responses and helical trajectory execution.
\end{enumerate}


The goal of this paper is to compare our aerial platform's behaviour with different SLAM algorithms being used for UAV pose estimation. To investigate this we will compare the UAV step responses for each case based on several integral criteria. Furthermore, the quality of trajectory execution is tested by comparing the RMS error based on the Hausdorff distance for an upward helical trajectory.

\subsubsection{Contributions}
Within this paper the emphasis is on the comparison of different 3D LiDAR SLAM algorithms based on their performance as a UAV pose sensor. The detailed overview of the system is presented, together with the fusion of the SLAM pose output with the IMU sensor through a Discrete Kalman Filter (DKF). Furthermore, the experimental analysis is conducted and the obtained results are presented and compared.

\subsubsection{Organization}
This paper is organized as follows. First the available SLAM algorithms and their intended usage is discussed in Section \ref{sec:slam}. Next, the detailed overview of the system is presented in Section \ref{sec:system_overview}. In Section \ref{sec:experiments} a detailed experimental analysis is conducted. Finally, the paper is concluded in Section \ref{sec:conclusion}.



\section{SLAM software selection} \label{sec:slam}

To find software candidates for the use of SLAM as a pose sensor, a GitHub search was conducted with the following criteria in mind:

\begin{enumerate}[label=(\roman*)]
    \item The software should be compatible with ROS Melodic and Ubuntu 18.04;
    \item The software should be actively maintained, well documented and easy to incorporate into our control system;
    \item The software package should have a scientific paper describing the proposed algorithm.
\end{enumerate}

The search returned 294 results, many of which were based on the LOAM algorithm \cite{Zhang_2014}. These were not taken into consideration because they either featured a version of LOAM optimised for ground vehicles such as \cite{legoloam2018}, or did not have a peer-reviewed paper describing their algorithm. From this software family we have chosen \cite{ALOAM_code}, an open source implementation of LOAM which we have tuned to be compatible with our use case.

Other interesting results include Berkeley Localization and Mapping \cite{BLAM} and GTSAM \cite{GTSAM}. These repositories were not included in our comparison because neither of them is backed by a publication. Furthermore, GTSAM can not be run with ROS without writing additional code. BLAM on the other hand fully supports ROS Kinetic, but the last commit to the repository has been made in 2016, so we consider it an inactive repository.

In the end, the software selection process has resulted with three SLAM systems: the Cartographer graph SLAM method \cite{Hess_2016}, LiDAR odometry and mapping in real time through LOAM \cite{Zhang_2014}, and the HDL graph SLAM method \cite{Koide_2019}. The remainder of this section gives a brief introduction to these algorithms. Its goal is to give an overview of the algorithms' general working principles, advantages and disadvantages, rather than their underlying mathematics.

\subsection{LOAM}

This algorithm does not rely on any other data besides the LiDAR sweeps to generate a map. Hence the name, LiDAR Odometry And Mapping. The algorithm extracts feature points from the received LiDAR point clouds and tracks their relative motion to estimate the sensor odometry. The map is a single point cloud stored as a KD-Tree, which is updated at a lower frequency. The algorithm is much simpler than Cartographer or HDL\_graph\_SLAM and has less parameters for tuning.

Unlike the other SLAM algorithms considered in this work, LOAM does not support loop closures. This means that it cannot recognise a previously mapped area and use this information to correct the pose estimate.

\subsection{Cartographer}
Cartographer is a graph SLAM method which uses IMU and LiDAR data. The IMU data is used to produce an initial estimate of the robot motion, while the LiDAR data is used to correct the IMU based estimate through scan matching against the built map. The map is divided into many smaller sub-maps which are created based on the IMU and LiDAR data in a process called local SLAM. The sub-maps are periodically rearranged to reduce odometry errors in a process called global SLAM. The global SLAM phase can use scan matching to detect loop closures, as well as the GPS data to refine the estimated robot trajectory. 

\subsection{HDL\_graph\_SLAM}

This algorithm is more recent than Cartographer and LOAM. It was presented in \cite{Koide_2019} where it was used on a backpack mapping system. Like Cartographer, it is a graph SLAM method so the two algorithms share some similarities. For example, both methods build a pose graph which is refined with loop closures, and optionally, GPS data. The main difference is that, unlike Cartographer, this SLAM system does not use the IMU data for odometry estimation. Odometry is computed from the LiDAR data directly by computing the transformation required to overlap recent scans. HDL graph SLAM allows the use of IMU data  to periodically improve the estimated trajectory by aligning the IMU acceleration vector of each trajectory node with the gravity vector. Additionally, the magnetic sensor can be used to achieve the same effect using the Earth's magnetic field. These IMU-based trajectory corrections were not used in our experiments.

\section{System overview} \label{sec:system_overview}

Within this section hardware and software components of the aerial system used in this work are described . 

\begin{figure}[t]
	\centering
	\vspace*{0.2cm}
	\includegraphics[width=0.99\linewidth]{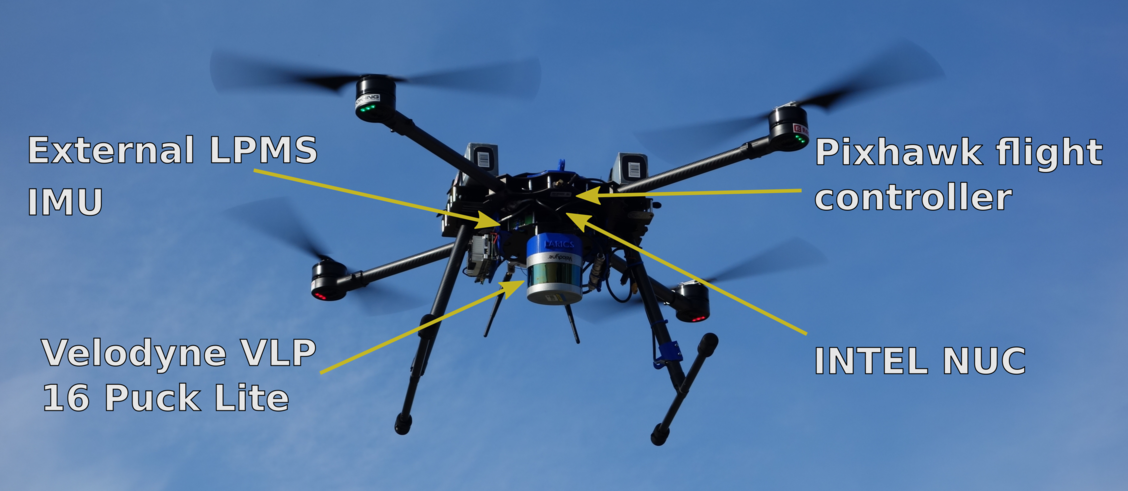}
	\caption{The Kopterworx quadcopter with the necessary sensors for performing SLAM.}
	\label{fig:hardware}
\end{figure}

\subsection{Hardware}
The UAV used is a custom quadcopter designed and assembled by the \textit{Kopterworx} company, depicted on Fig. \ref{fig:hardware}. The UAV features a lightweight carbon fiber body with a propulsion unit mounted on each quadcopter arm. The propulsion unit consists of the \textit{T-Motor P-60 170KV} brushless  motor and folding \textit{T-Motor MF2211} $22.4in \times 8.0^\circ$ propeller. The maximum thrust of this particular motor-propeller pair reaches $68N$. The UAV is powered by two \textit{Tattu 14000mAh 6S} batteries which allow a total flight time of around $30min$. Furthermore, we use the \textit{Pixhawk 2.1} autopilot running the \textit{Ardupilot} flight stack. The pilot can operate the UAV using the \textit{Futaba T10J} transmitter, operating on the $2.4GHz$ band and paired with a \textit{Futaba R3008SB} receiver which is mounted on the vehicle. Optionally, the telemetry can be monitored through a separate $868MHz$ channel.

To run computationally expensive high-level SLAM algorithms in real time, an \textit{Intel NUC} onboard computer is mounted on the UAV. It features an \textit{Intel Core i7} processor and runs \textit{Linux Ubuntu 18.04} with Robot Operating System - \textit{ROS Melodic} installed. The communication between the onboard computer and the flight controller is realized through a serial port. The Linux-based onboard computer supports interfacing a broad sensory apparatus. For the purposes of this paper, the UAV is equipped with a \textit{Velodyne Puck} LiDAR sensor for the point cloud acquisition, and with an external \textit{LPMS-CU2} IMU capable of providing high rate measurements. The LiDAR has a thirty degree vertical field of view and a vertical resolution of 1.8 degrees. It is operated at 1200 rpm which gives it a horizontal resolution of 0.4 degrees.

\subsection{Software}
\begin{figure}[t]
	\centering
	\vspace*{0.2cm}
	\includegraphics[width=\linewidth]{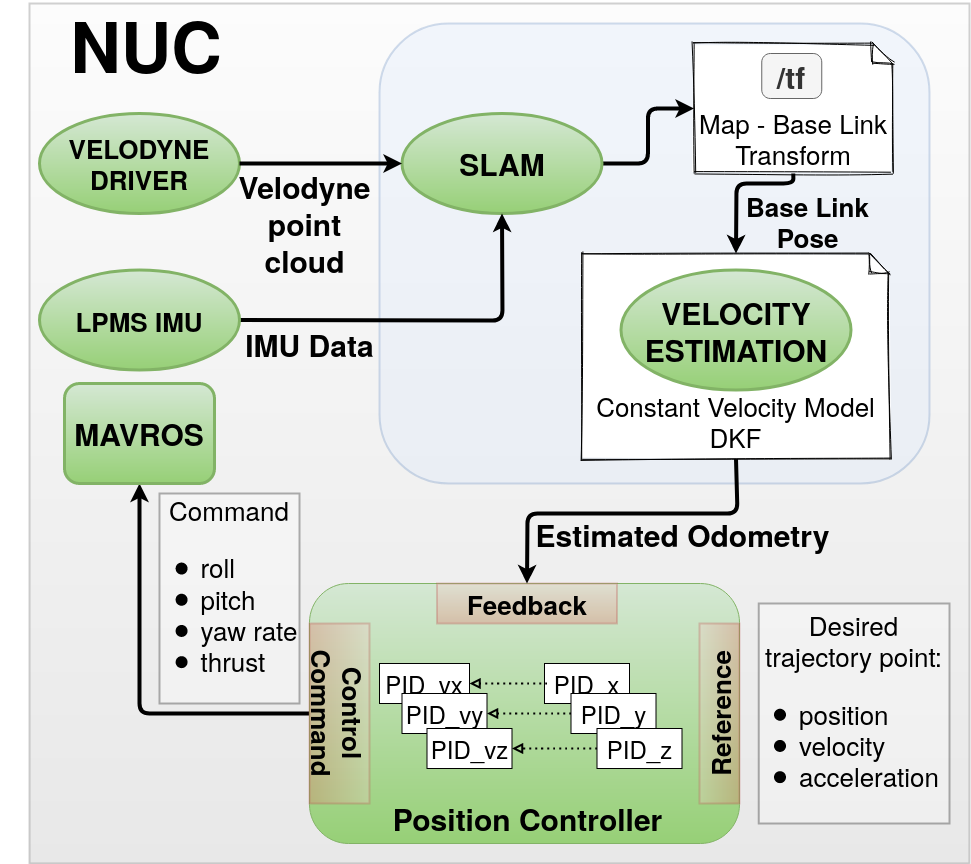}
	\caption{Diagram of ROS nodes representing the estimation and control stack used in the experiments.}
	\label{fig:ros_stack}
\end{figure}

The software on the UAV is running on both the flight controller and the onboard computer, as depicted on Fig. \ref{fig:ros_stack}. The \textit{Pixhawk} flight controller is mainly responsible for the low-level attitude control, interfacing the radio communication and telemetry, and fusing the raw sensor data for the attitude controller feedback. 

On the other hand, the onboard computer is capable of running computationally expensive tasks. This work focuses on three components: high-level position and velocity control; communication between the onboard computer and the flight controller; and using SLAM algorithms for obtaining reliable position and velocity feedback.

\subsubsection{Position and velocity control}
\label{sec:pos_vel_control}
To control the UAV a standard PID cascade is employed with the inner loop controlling the velocity and the outer loop controlling the position. The reference for this controller can be a position setpoint or a smooth trajectory. The latter typically contains additional velocity and acceleration setpoints. To account for these dynamic values, feed forward terms are added to the controller. 

Since the controller is unaware of the feedback source, GPS can be used in outdoor environments, or motion capture systems such as \textit{Optitrack} for indoor laboratory experiments. SLAM algorithms that can localize the UAV only using the onboard sensory apparatus can, on the other hand, be used in both scenarios. Based on the reference and the feedback, the controller produces output values to the flight controller. In our case, these values are roll angle, pitch angle, yaw angular rate and thrust, which are the inputs to the attitude controller.

\subsubsection{Communication}
As mentioned earlier, the communication between the flight controller and the onboard computer is carried out through a serial port. The protocol in use is the widely accepted Micro Air Vehicle Link \textit{MAVLink}, which is supported by the \textit{Ardupilot} flight stack. The onboard computer relies on the open source \textit{mavros} package that acts as a ROS wrapper for the \textit{MAVLink} protocol. Through \textit{mavros} we are receiving all relevant data from the \textit{Pixhawk}, namely: IMU data; radio controller switch positions; GPS data. As the communication is two-way, we are also sending the attitude controller setpoint that is computed by the position controller.

\subsubsection{Feedback}
As mentioned in Section \ref{sec:pos_vel_control}, the feedback can be obtained through various channels. The focus of this paper is on the usage of SLAM algorithms for feedback. The algorithms considered in this paper operate on the point cloud data, which is obtained through the \textit{Velodyne Puck} LiDAR sensor. The output is position and orientation of the UAV in the inertial coordinate system. Since our controller expects velocity as well as position measurements, we employ a Discrete Kalman Filter (DKF) for the velocity estimation. The estimated position and velocity outputs from the DKF are used as feedback for the controller.

The prediction step of the DKF uses a constant acceleration model given as follows:
\begin{gather}
	\boldsymbol{\zeta}_{k+1} = \text{F}_k  \boldsymbol{\zeta}_k + \boldsymbol{\omega}_k \, , \\
	\qquad \text{F}_k =     
	\begin{bmatrix}
	\text{A}_k & \text{0}_{3 \times 3} & \text{0}_{3 \times 3} \\
	\text{0}_{3 \times 3} & \text{A}_k & \text{0}_{3 \times 3} \\
	\text{0}_{3 \times 3} & \text{0}_{3 \times 3} & \text{A}_k
	\end{bmatrix}          
	\, , \\ 
	\text{A}_k =     
	\begin{bmatrix}
	1 & T_s & \frac{T_s^2}{2} \\
	0 & 1 & T_s \\
	0 & 0 & 1
	\end{bmatrix}          
	\, , 
\end{gather}
where $\text{F}_k \in \mathbb{R}^{9 \times 9}$ is the model transition matrix, $\boldsymbol{\omega}_k \in \mathbb{R}^9$ represents process noise for the respective states which are defined as follows:
\begin{equation}
    \boldsymbol{\zeta}_k = \left[ x, \dot{x}, \ddot{x}, y, \dot{y}, \ddot{y}, x, \dot{z}, \ddot{z}  \right]^{\text{T}} \, .
\end{equation}
The available measurements used are the position provided by the SLAM algorithm and the acceleration obtained from the \textit{LPMS CU2} IMU sensor. Therefore the observation model is given as follows:
\begin{gather}
    \textbf{z}_k = \text{H}_k \boldsymbol{\zeta}_k + \boldsymbol{\upsilon}_k \, , \\
    \text{H}_k = 
	\begin{bmatrix}
	\text{B} & \text{0}_{3 \times 3} & \text{0}_{3 \times 3} \\
	\text{0}_{3 \times 3} & \text{B} & \text{0}_{3 \times 3} \\
	\text{0}_{3 \times 3} & \text{0}_{3 \times 3} & \text{B}
	\end{bmatrix} \, , \\
	\text{B} =     
	\begin{bmatrix}
	1 & 0 & 0 \\
	0 & 0 & 0 \\
	0 & 0 & 1
	\end{bmatrix} \, ,
\end{gather}
where $\text{H}_k \in \mathbb{R}^{9 \times 9}$ is the observation transition matrix, $\boldsymbol{\upsilon}_k \in \mathbb{R}^9$ represents the measurement noise. Both $\boldsymbol{\omega}$ and $\boldsymbol{\upsilon}$ are independent zero-mean Gaussian distribution vectors with respective correlation matrices $\text{Q}$ and $\text{R} \in \mathbb{R}^{9 \times 9}$. Diagonal components of the $\text{R}$ matrix are obtained through signal-to-noise ratios of position and acceleration measurements. Diagonal components of the $\text{Q}$ matrix are tuned to minimize estimated position and velocity noise and delay. 
The remainder of the DKF correction update equations are omitted for brevity. 
\section{Experiments} \label{sec:experiments}

In order to compare the SLAM methods, two sets of experiments were conducted. The first set was used as a safety measure to test the SLAM algorithms offline on three datasets  before using the SLAM algorithms for pose estimation in our control system. These datasets were designed to test different aspects of the algorithms. The first, baseline dataset, was recorded while flying in a circle around a small hangar. That dataset features mild trajectories and is expected not to cause problems for SLAM algorithms. In contrast, the second dataset features dynamically challenging UAV trajectories in an attempt to destabilize the SLAM systems. Finally, the third dataset featured a 12 minute flight with non-agile trajectories to compare the SLAM methods on a large mapping task. In all of these experiments the SLAM feedback quality is compared based on the algorithms' estimated positions and orientations. Furthermore, a coarse comparison of the SLAM systems' precision is achieved by comparing the reported takeoff and landing locations with the ground truth registered by the \textit{OptiTrack} motion capture system installed in the hangar. These results are reported in table \ref{table:results_landing_coordinates}.

After the offline SLAM comparison, we integrated the SLAM methods into the UAV control system to serve as a pose sensor. The control system quality is assessed using step excitation signals and helical trajectories. For the step responses various integral performance criteria are used for the comparison, together with the percentage of overshoot and the rise time. For the trajectory following experiments, the RMS error based on the Hausdorff distance is used. In these experiments, the UAV is first flown manually to create an initial map of the environment before switching to the control system and generating the required signals.

\subsection{Offline SLAM comparison}

As expected, the SLAM algorithms didn't have any problems on the first dataset. Their reported trajectories are similar, and so are their built maps.

\begin{figure}[t]
	\centering
	\vspace*{0.2cm}
	\includegraphics[trim={0.5cm 0 1cm 1.5cm},clip,width=0.99\linewidth]{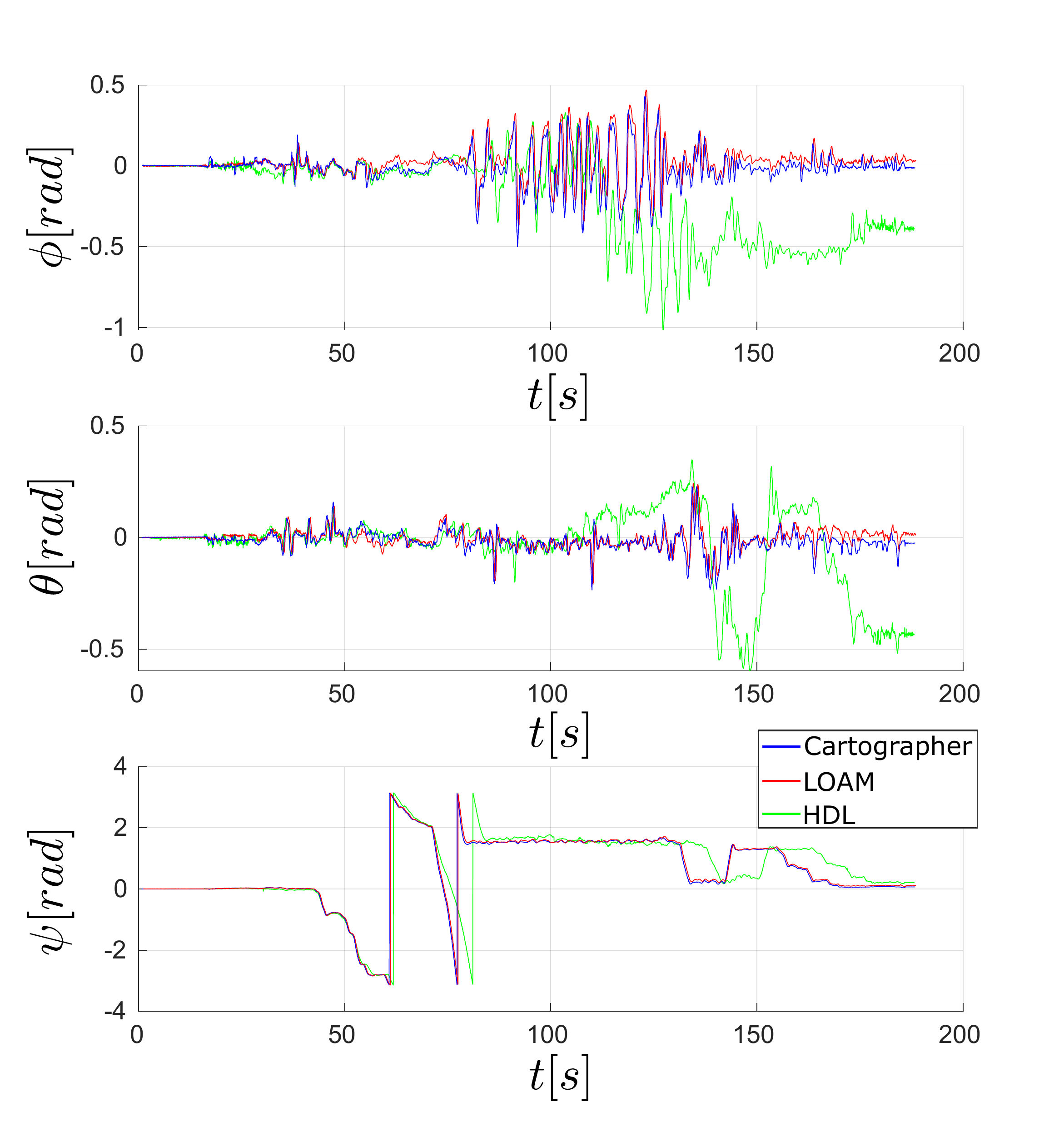}
	\caption[Roll pitch and yaw angles]{Roll, pitch and yaw angles reported by the SLAM algorithms for the second dataset. It is visible that the roll angle reached up to 0.5 radians. The maximum allowed roll angle of our UAV is 0.8 radians.}
	\label{fig:preliminary2angles}
\end{figure}

The second dataset proved to be more challenging because of sudden high-amplitude UAV angle changes. Fig. \ref{fig:preliminary2angles} displays those angles. Therefore, the SLAM algorithms were tuned specifically for this dataset, and these tunings are used in the remainder of this paper. 

Despite our best efforts, we were unable to tune HDL\_graph\_SLAM to produce a coherent map on this dataset. For this reason, we have excluded this algorithm from our comparison.

\begin{figure}[t]
	\centering
	\vspace*{0.2cm}
	\includegraphics[trim={1.0cm 0 2.5cm 2.0cm},clip,width=0.99\linewidth]{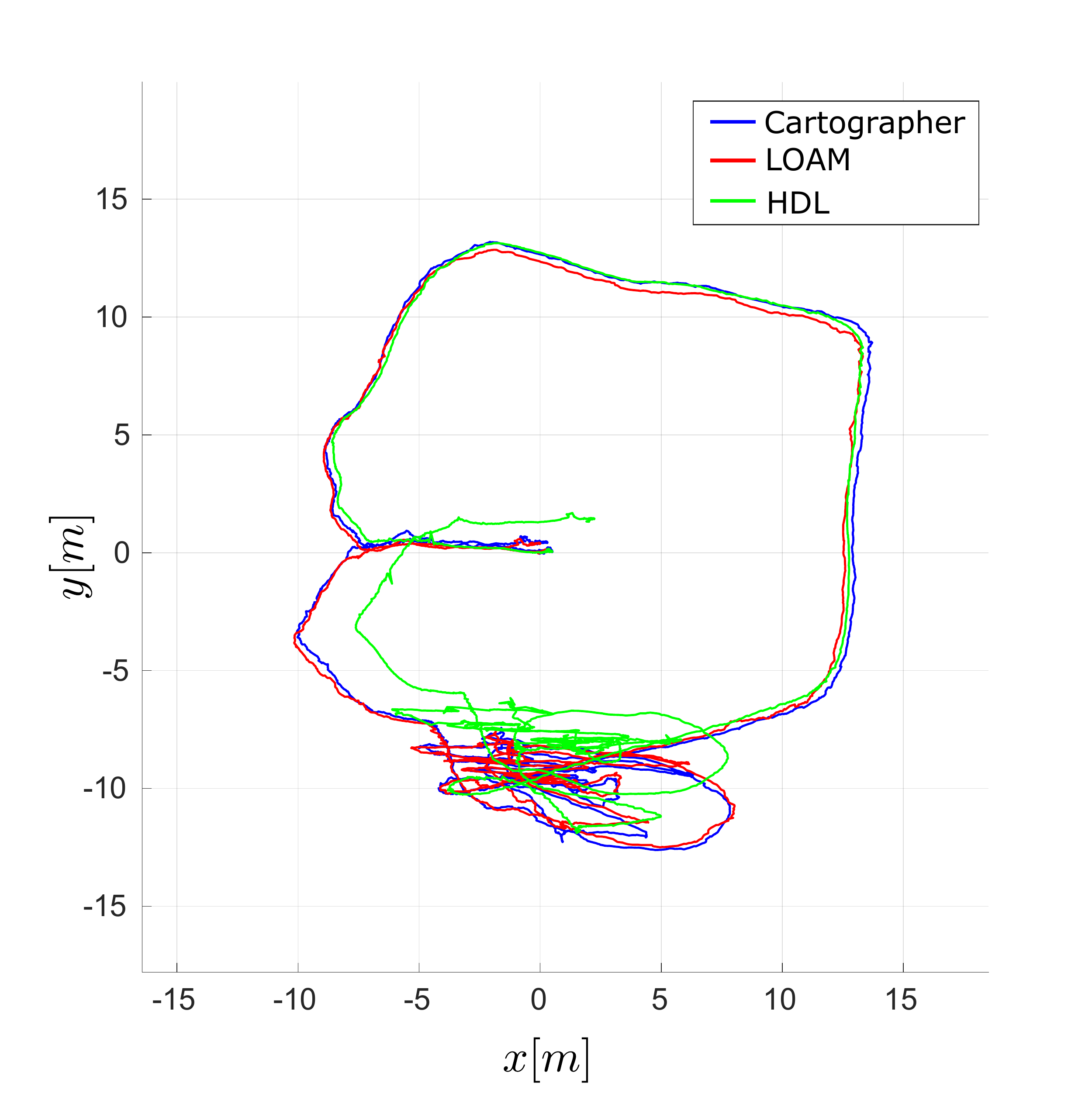}
	\caption{Comparative x-y plot for the second dataset. Cartographer and LOAM report similar x-y positions.}
	\label{fig:preliminary_2_xy}
\end{figure}

The third dataset really shed light on the differences of the SLAM algorithms since it was collected around a long building on the UNIZG campus (Fig. \ref{fig:preliminary_3_carto_points}). 

\begin{figure}[th]
	\centering
	\vspace*{0.2cm}
	\includegraphics[trim={0.5cm 0 2.0cm 1.5cm},clip,width=0.99\linewidth]{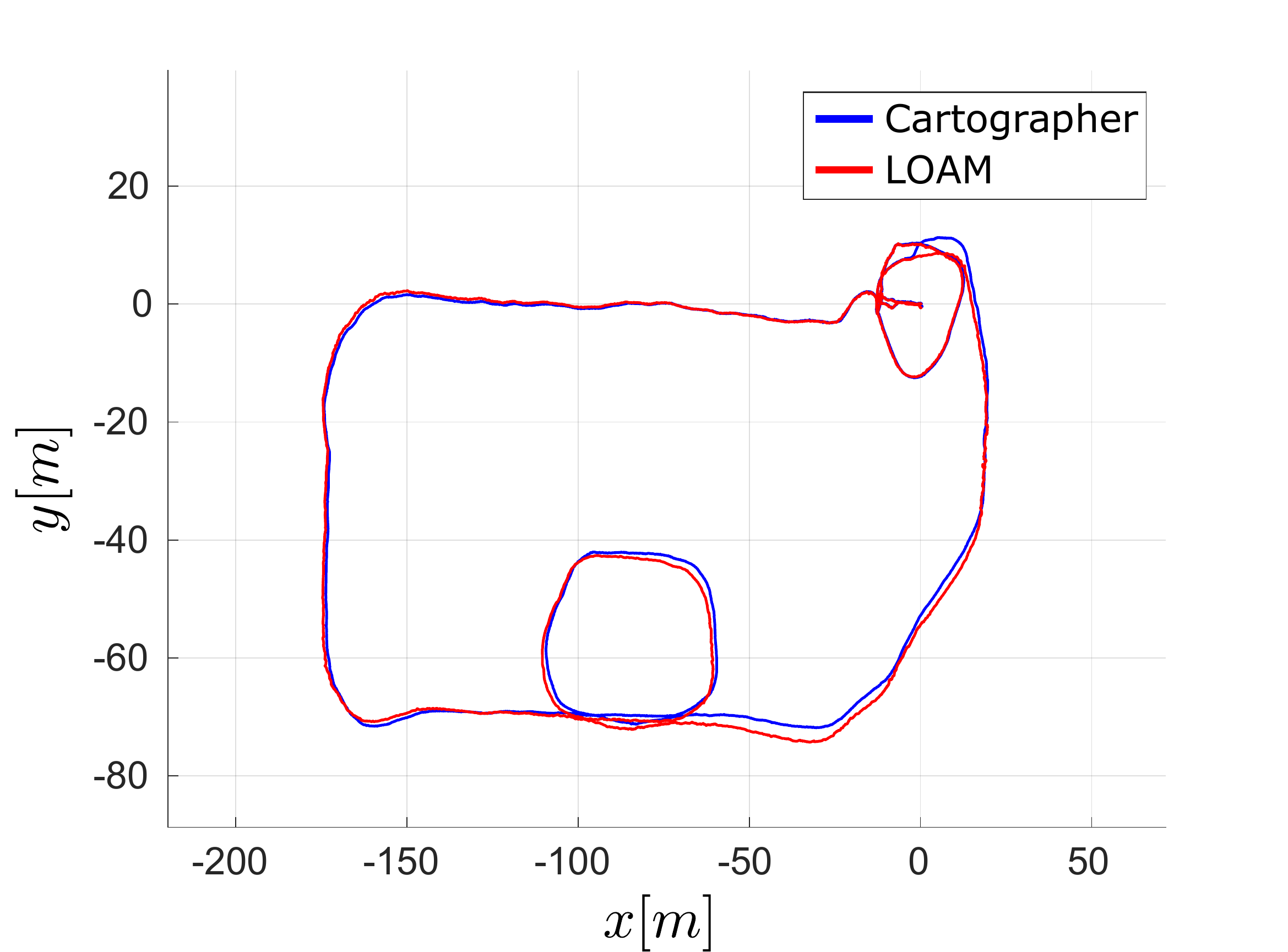}
	\caption{Comparative x-y plot for the third dataset. One can see how Cartographer's readings start to diverge from LOAM's after the lower loop. This happens because Cartographer's local SLAM phase accumulates more drift than LOAM. This can be seen by visually inspecting the SLAM process which is shown in the supplementary video material \cite{video_playlist}, where Cartographer's map reports two partially overlapping buildings in the upper right corner of the map. This error is later corrected by loop closing which causes the poses to converge.}
	\label{fig:preliminary_3_xy}
\end{figure}
\begin{figure}[th]
	\centering
	\vspace*{0.2cm}
	\includegraphics[trim={0.2cm 0 1.9cm 1.5cm},clip,width=0.99\linewidth]{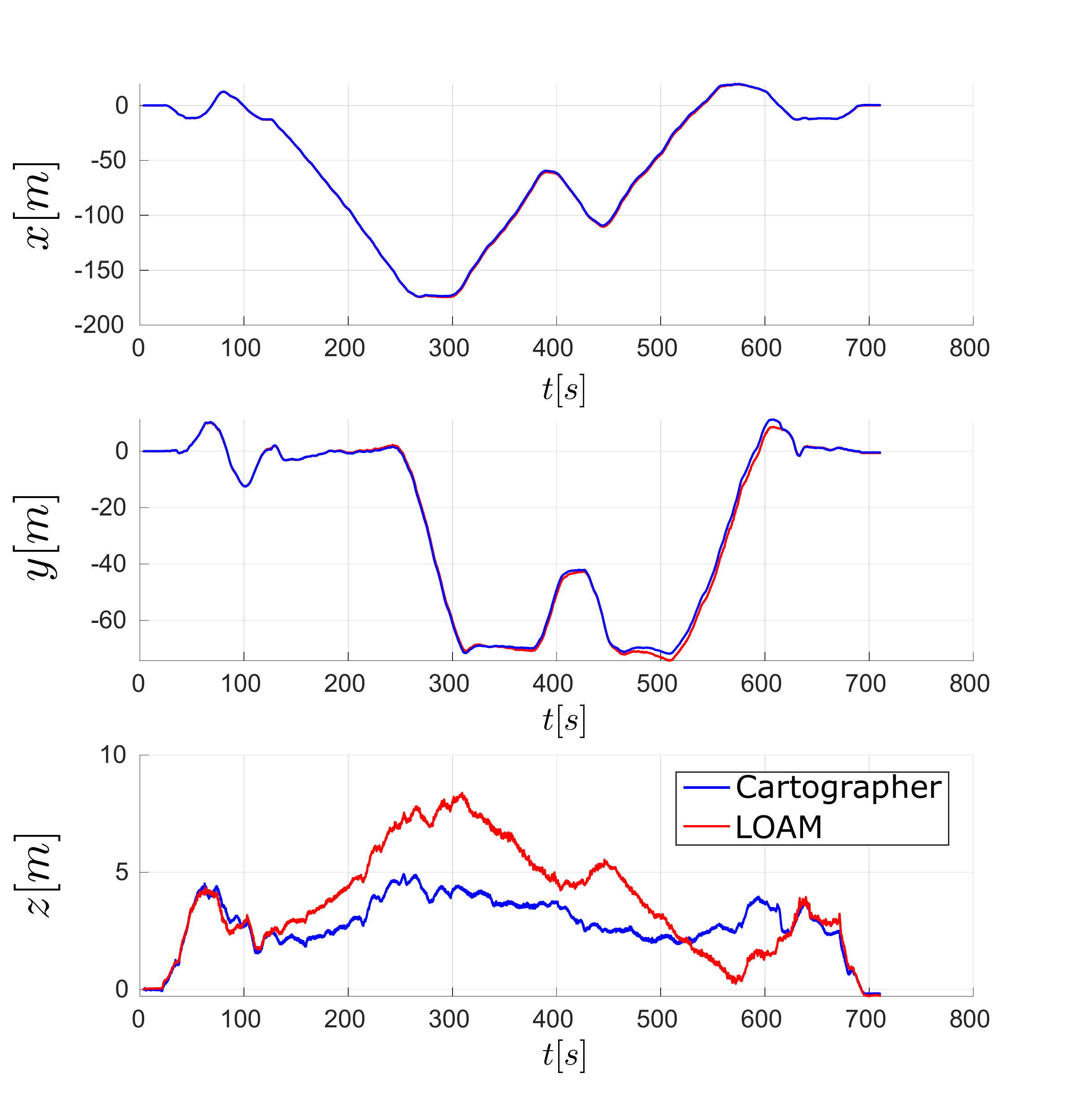}
	\caption{Estimated UAV position for each coordinate axis. The z axis plot shows a drastic difference between LOAM and Cartographer. Since LOAM computes LiDAR only odometry, a significant amount of drift was introduced in the middle part of the flight. Since cartographer uses an IMU sensor for prior pose estimation, its z axis measurement is more accurate. This can be stated despite not having a high precision ground truth because it is known that the ground on the mapping location is level and we have not flown higher than 4 meters during the manual data recording.
	Also, this graph shows how Cartographer started to diverge from LOAM in the y axis after the 450th second. A pose difference visible to the naked eye persists until Cartographer manages to close a loop and correct the accumulated drift. That moment is visible in the y dimension after the 600th second where the poses converge once more.}
	\label{fig:preliminary_3_coordinates}
\end{figure}

This site was chosen in order to stimulate the SLAM systems to accumulate  drift. It can be observed that LOAM accumulates less drift than Cartographer, but since LOAM cannot detect loop closures on its own, that drift never gets corrected. On the other hand, Cartographer accumulates a significant amount of drift, but upon returning to the vicinity of the takeoff position, that drift is corrected by loop closing through global SLAM. 
This can be seen on Figs. \ref{fig:preliminary_3_xy} and \ref{fig:preliminary_3_coordinates} and on the supplementary video material \cite{video_playlist}. 
The video shows that Cartographer creates two partially overlapped buildings in the upper right corner of the map. The two buildings are merged into one when the loop closure happens.



Apart form the loop closure, Fig.  \ref{fig:preliminary_3_coordinates} also shows that Cartographer and LOAM report different altitudes. Since the LiDAR sensor has a vertical angle of only 30 degrees, the sweeps contain more information about the x and y axes than about the z axis. This causes the z coordinate to have a greater uncertainty than x and y. 
Cartographer manages to reduce the drift in the z dimension by using an IMU for prior pose estimation.

Since our goal is to use SLAM as a pose sensor, loop closures pose a threat to the control system because they introduce step signals into the pose measurements. To tackle this problem an exponential filter is incorporated into Cartographer which smooths the steps across a five second period.

\begin{table}[h]
	\caption{Absolute error of landing coordinates reported by SLAM on the three datasets}
	\label{table:results_landing_coordinates}
	\centering
	\begin{tabular}{cccc} 
	\hline
	\multicolumn{4}{c}{Absolute error at landing position} \\
	\hline \hline
	LOAM & 0.26 m & 0.51 m & 0.52 m \\ 
	Cartographer & 0.27 m & 0.61 m & 0.23 m \\ 
	\hline 
	\end{tabular}
\end{table}

\subsection{Online comparison}

For the second set of experiments the SLAM methods are used as feedback to the control system and the UAV is set to follow a generated reference signal. The experiments were carried out on a calm day to minimize the effect of unknown disturbances, such as wind. Before each experiment the UAV is flown manually with the SLAM algorithm running to generate a map of the environment. Cartographer pose measurements arrive at 50 Hz, while LOAM pose measurements have a rate of 20 Hz. This difference occurs because Cartographer's rate depends on the IMU frequency, while LOAM's rate depends on the LiDAR frequency.

\subsubsection{Step responses}

\begin{figure}[th]
	\centering
	\vspace*{0.2cm}
	\includegraphics[trim={0.7cm 0 1.9cm 1.5cm},clip,width=0.99\linewidth]{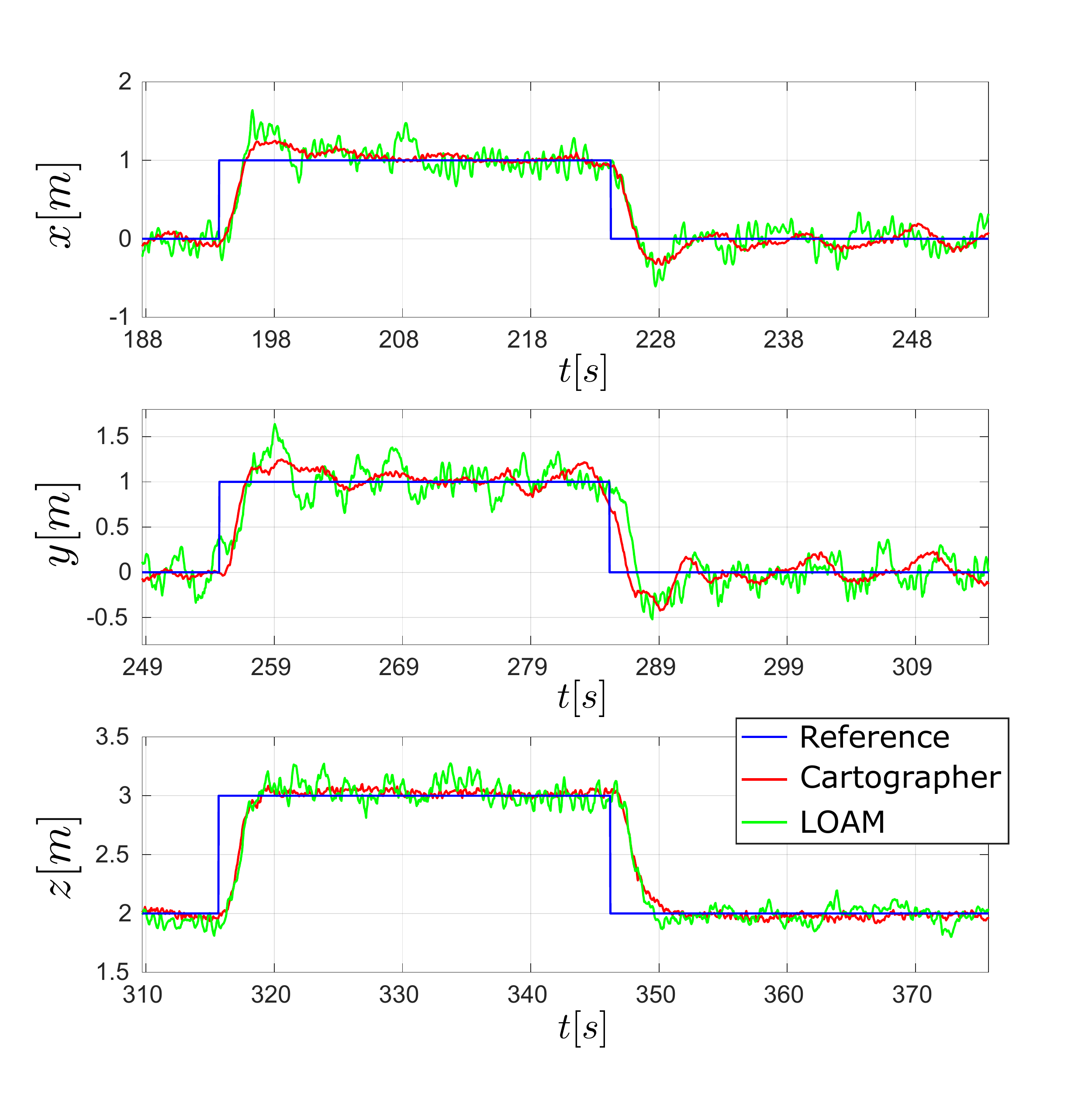}
	\caption{Comparison of step responses with Cartographer and LOAM as pose sensor. The pose feedback is more oscillatory when it is obtained through LOAM.}
	\label{fig:steps}
\end{figure}

To record the step responses the UAV is flown in the middle of the generated map  and position hold mode is activated.
With the control system active, a series of positive and negative step signals is generated for each axis separately. The responses are displayed on Fig. \ref{fig:steps}. One can see that the UAV oscillates more when the pose is supplied by LOAM. This is due to Cartographer having a faster pose publishing rate. 

To compare the step responses numerically the IAE, ISE, ITAE and ITSE integral performance criteria were calculated for each step response. These values are displayed in the Table \ref{table:integral_criteria}. In addition to these criteria, the overshoot and rise time values are shown in the same table.

\begin{table}[b]
		\caption{Various integral performance criteria, percentage of overshoot $PO$, and rise time $t_r$ calculated from the recorded step responses.}
	\label{table:integral_criteria}
	\centering
	\begin{tabular}{c|c|cccc|cc}
		\hline
		\multicolumn{2}{c|}{ } & IAE & ISE & ITAE & ITSE & $PO[\%]$ & $t_r[s]$  \\ 
		\hline
		\parbox[t]{2mm}{\multirow{3}{*}{\rotatebox[origin=c]{90}{Carto}}} & x & 336 & 3031 & 8175 & 90660 & 25 & 1.2 \\ 
		& y & 357 & 3025 & 8940 & 89680 & 25 & 1.0 \\ 
		& z & 252 & 3012 & 5368 & 88788 & 5 & 1.6 \\ 
		\hline
		\parbox[t]{2mm}{\multirow{3}{*}{\rotatebox[origin=c]{90}{LOAM}}} & x & 511 & 3023 & 12554 & 90039 & 60 & 1.3\\ 
		& y & 521 & 3029 & 13204 & 90238 & 60 & 2.0 \\ 
		& z & 362 & 3005 & 8176 & 88555 & 20 & 1.6\\ 
		\hline 
	\end{tabular} 
\end{table}



\subsubsection{Trajectory following}

To test the control system on a dynamic trajectory, two upward helical trajectories were generated. The slower trajectory has velocity constraints of $v_s = 1m/s$ and acceleration constraints of $a_s = 0.5m/s$ for all axes. The faster trajectory has two times higher constraints than the slower one. On lower altitudes the LiDAR can detect a lot of nearby features such as walls and rooftops. Flying on higher altitudes poses a greater challenge for LiDAR SLAM because the number of detectable features decreases resulting in sparser point clouds. This causes the LOAM pose estimate quality to deteriorate and ultimately, our safety pilot has to take control of the UAV. Cartographer doesn't have this problem at that altitude since it relies on IMU measurements to form a prior pose estimate.

\begin{figure}[th!]
	\centering
	\includegraphics[trim={1.0cm 0 1.9cm 1.5cm},clip,width=0.99\linewidth]{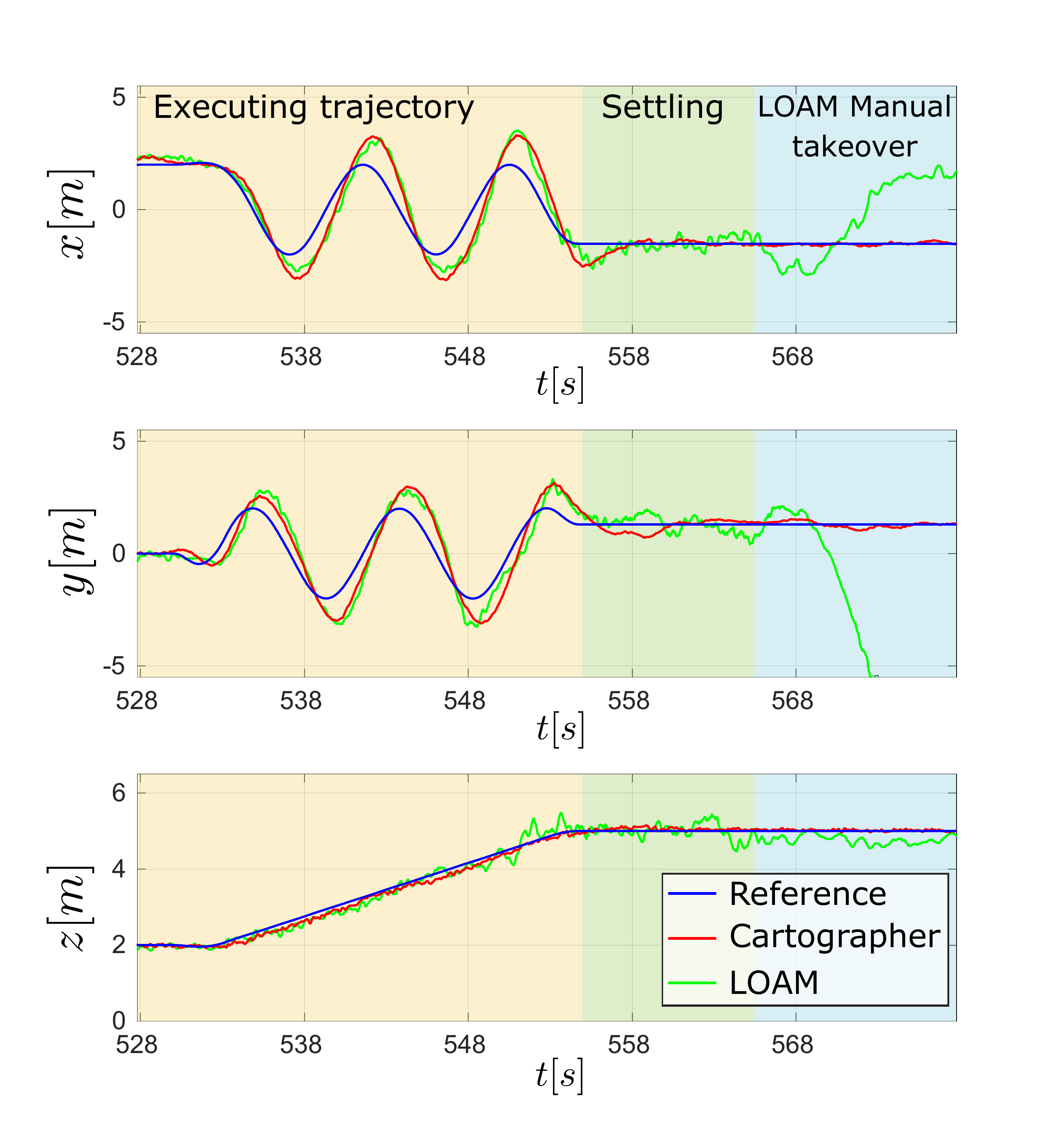}
	\caption{Comparison of helical trajectory execution for each axis. The systems track the trajectories similarly until the altitude of 4 m is reached. At that moment LOAM feedback quality starts to drop as the LiDAR slowly looses sight of the nearby building. Ultimately, our safety pilot had to take control of the UAV.}
	\label{fig:fast_helix}
\end{figure}

Fig. \ref{fig:fast_helix} compares the trajectory execution for each axis. Until the altitude of 4m is reached, the systems track the trajectories similarly except for the small amplitude oscillations present in the LOAM feedback. To numerically compare the responses, the RMS error based on the Hausdorff distance between the planned and the executed trajectory was employed. The finally obtained RMS for Cartographer is $RMS_c = 0.6365m$ for slower and $RMS_c = 0.9194m$ for faster trajectory. The values obtained for the LOAM are $RMS_l = 0.6543m$ for slower and $RMS_l = 0.8520m$ for faster trajectories. These were computed based on the responses until the moment of manual takeover. The computed values are similar for the two systems, therefore Cartographer's advantage in this scenario is the ability to fly at higher altitudes.

\section{Conclusion} \label{sec:conclusion}

After comparing the selected SLAM algorithms, it can be concluded that LOAM introduces less drift into the system than Cartographer, but when used in a UAV control system for pose feedback Cartographer outperforms LOAM. Since the state of the art uses LOAM fused with other sensors for this purpose, it would be beneficial to compare LOAM and Cartographer in that setting in some future work.

However, based on the experiments conducted in this work, it can be concluded that Cartographer is a better choice for a light-weight aerial pose sensing application.
The fact that Cartographer already relies on an IMU for prior pose estimation gives it an advantage in situations where the environment is poor in LiDAR features, such as higher altitude flights. Furthermore, the loop closing feature allows it to correct the accumulated drift allowing higher pose accuracy to be achieved during long flights.

The video accompanying the experimental analysis in this paper can be found on YouTube \cite{video_playlist}.

\section*{ACKNOWLEDGMENT}

This work has been supported by the European Commission Horizon 2020 Programme through project under G. A. number 810321, named Twinning coordination action for spreading excellence in Aerial Robotics - AeRoTwin \cite{AEROTWINweb} and through project under G. A. number 820434, named ENergy aware BIM Cloud Platform in a COst-effective Building REnovation Context - ENCORE \cite{ENCOREweb}.

\bibliographystyle{ieeetr}
\bibliography{bibliography/bibliography.bib}

\end{document}